\documentclass[]{arxiv_article}

\usepackage{multirow}
\microtypesetup{expansion=false}
\usepackage{xspace}
\usepackage{enumitem}
\usepackage[linesnumbered,ruled,vlined]{algorithm2e}
\usepackage{amsmath}
\usepackage{amssymb}
\definecolor{darkgreen}{rgb}{0.0,0.45,0.0}

\newcommand{\keywords}[1]{\par\medskip\noindent\textbf{Keywords:}\space\def\and{, }#1}

\title{DSWAM: A Dual-System World Action Foundation Model for Fine-Grained Robot Manipulation}

\author[1,*,\dagger,\ddagger]{Jian Zhu}{}
\author[1,2,*]{Jianjun Zhang}{}
\author[1,*]{Taiyi Su}{}
\author[1,*]{Tianbin Liu}{}
\author[1]{Zhangyuan Wang}{}
\author[1]{Kai Xie}{}
\author[1,2]{Zitai Huang}{}
\author[1,2]{Chong Ma}{}
\author[1]{Youzhang He}{}
\author[1]{Tianjian Wang}{}
\author[1]{Hanyang Wang}{}
\author[1]{Weihao Ding}{}
\author[1,\dagger]{Yi Xu}{}

\affiliation[1]{AIRC, Midea Group}
\affiliation[2]{Tongji University}

\contribution[*]{Equal Contribution}
\contribution[\dagger]{Corresponding Author}
\contribution[\ddagger]{Project Leader,}

\abstract{
World Action Models (WAMs) provide a promising alternative to Vision-Language-Action (VLA) policies by using video-based world modeling as dense supervision for robot action learning. Existing WAMs excel at physically grounded execution, but typically lack the explicit language-level planning interface in VLM-based VLAs for decomposing coarse instructions. Such decomposition becomes important when household tasks involve complex multi-step goals, where coarse user commands need to be converted into sequences of fine-grained executable subtasks. Meanwhile, the field still lacks a fair real-robot comparison between VLA and WAM execution capabilities, since existing systems often differ in data, robot embodiments, and task protocols.
To address both the decomposition gap and the need for a controlled WAM-VLA comparison, we introduce \textbf{DSWAM}, a Dual-System World Action Foundation Model for fine-grained robot manipulation. DSWAM keeps a System 1 WAM executor as the default control path and optionally activates a System 2 vision-language subtask planner only when task decomposition is useful. The planner predicts executable subtasks from short-term visual history and a global task prompt, while the WAM executor performs world-aware action generation for each instruction or subtask. The executor is trained with action prediction and video co-training, but inference directly predicts action chunks without explicit future video generation. To make this execution path practical on real robots, we further integrate TensorRT acceleration, asynchronous execution, and real-time chunking (RTC) so that policy queries do not block robot control.
To provide a fair real-robot comparison with VLA policies, we build and evaluate DSWAM under the DeMaVLA real-world deformable manipulation setting with matched robot platform, pretraining data, post-training data, and evaluation criteria. In this folding benchmark, DSWAM runs in WAM-only mode with the System 2 planner disabled. Under this setting, DSWAM improves average real-world folding success rate from 92.5\% to 96.3\% and reduces completion time from $2'18''$ to $1'44''$. We further show that the optional System 2 subtask supervision improves the real-world execution stability on tasks that benefit from decomposition by increasing success rate and reducing rollout mistakes.

\keywords{World Action Model, Dual-System, Fine-Grained Robot Manipulation, Video Co-training, TensorRT Acceleration}
}

\date{\today}
\correspondence{Jian Zhu, Yi Xu}
\checkdata[Project Page]{\url{https://ds-wam.github.io/}}

\begin{document}
\begingroup
\hypersetup{linkcolor=black}%
\maketitle
\endgroup

\section{Introduction}
\label{sec:intro}

World Action Models (WAMs) have recently emerged as a promising alternative to Vision-Language-Action (VLA) policies for robotic manipulation~\citep{ye2026world,kim2026cosmos,bi2025motus,li2026causal,yuan2026fastwam}. Existing VLA policies convert visual observations and language instructions into executable robot actions~\citep{kim2024openvla,black2024pi_0,intelligence2025pi,pertsch2025fast,cheang2025gr,bjorck2025gr00t}, but their common single-frame, observation-to-action formulation provides limited temporal context for modeling scene evolution during robot manipulation. WAMs address this limitation by learning how visual states evolve under actions, providing world-aware representations of physical dynamics. This video-based formulation gives dense supervision for robot action learning and enables strong performance in contact-rich manipulation tasks such as grasping, placing, and deformable object manipulation, where accurate modeling of physical interactions is critical.

However, when household tasks involve complex multi-step goals, physically grounded execution alone is not enough. Coarse user commands often need to be converted into sequences of fine-grained executable subtasks. VLM-based VLAs provide a natural language-level interface for semantic grounding, planning, and instruction decomposition, whereas existing WAMs are mainly designed as execution policies and typically lack this capability. This suggests that semantic task organization and world-aware physical execution can be decoupled: a WAM can remain a strong execution engine, while a separate high-level module supplies task decomposition only when the instruction benefits from it. This avoids making planning mandatory for atomic instructions or tasks that the WAM executor already solves reliably, while still supporting coarse commands that require structured subtask execution. Beyond this modeling issue, real-robot comparisons between VLA and WAM policies are often confounded by differences in data sources, robot embodiments, and task protocols, making it unclear whether performance differences come from the policy design or the experimental setup.

To address these issues, we present \textbf{DSWAM}, a Dual-System World Action Foundation Model for fine-grained robot manipulation. DSWAM instantiates the above decoupling by preserving a System 1 WAM executor as the default control path and adding a System 2 vision-language subtask planner only as an optional high-level module. For atomic instructions, or for tasks that the WAM executor already solves reliably, DSWAM runs in WAM-only mode without invoking System 2. For coarse commands that benefit from decomposition, System 2 predicts an ordered sequence of executable subtasks from short-term visual history and the global task prompt, and System 1 executes each instruction or subtask through world-aware action generation. This design lets the high-level module focus on semantic task organization while leaving contact-rich physical execution to the WAM executor. The planner is trained with transition-aware subtask supervision, while the executor is trained with action prediction and video co-training. Following recent evidence that world-modeling supervision can be separated from test-time future imagination~\citep{yuan2026fastwam,ye2026gigaworld}, the executor directly predicts action chunks at inference time without explicit future video generation. To make this execution path practical on real robots, we combine real-time chunking (RTC)~\citep{black2026real}, asynchronous execution, and TensorRT acceleration so that policy queries do not block robot control. To provide a fair real-robot comparison between WAM-style execution and VLA policies, we build DSWAM under the DeMaVLA setting: the executor is pretrained on the same large-scale real-robot data to strengthen general manipulation capability, and evaluation follows the same robot platform, post-training data, task protocol, and success criteria~\citep{su2026demavla}. In this folding benchmark, DSWAM uses the WAM executor directly and does not start the System 2 planner, so the matched comparison focuses on policy execution capability rather than additional high-level planning. This controlled folding study shows that the WAM executor can outperform strong VLA references under the same real-robot setup, raising average success from 92.5\% to 96.3\% and shortening completion time from $2'18''$ to $1'44''$ without planner assistance. Beyond deformable manipulation, DSWAM reaches 92.38\% average success on RoboTwin 2.0 clean tasks and 91.90\% under randomized conditions, indicating broad simulated manipulation capability. Finally, our System 2 study shows that subtask-level supervision improves stability when a coarse instruction benefits from decomposition.

\textbf{We summarize our contributions as follows:}
\begin{itemize}
    \item We propose \textbf{DSWAM}, a dual-system WAM framework that decouples semantic task decomposition from world-aware physical execution: System 1 serves as the default WAM executor, while an optional System 2 vision-language subtask planner supplies ordered executable subtasks only when decomposition is beneficial.
    \item We build a scalable real-robot WAM foundation policy under the matched DeMaVLA setting, using the same robot platform, pretraining data, post-training data, task protocol, and evaluation criteria, and run the folding benchmark with System 2 disabled to fairly compare policy execution capability with VLA policies.
    \item We report strong empirical results: DSWAM improves matched DeMaVLA real-world folding success from 92.5\% to 96.3\%, reduces completion time from $2'18''$ to $1'44''$, reaches 92.38\% and 91.90\% success rate on RoboTwin 2.0 clean and randomized tasks, and shows that optional System 2 subtask supervision improves execution stability when decomposition is useful.
\end{itemize}

\section{Related Work}
\label{sec:related}

\subsection{Vision-Language-Action Policies}

Vision-Language-Action (VLA) policies connect visual observations, language instructions, and robot actions in a unified policy model. By building on pretrained vision-language backbones, recent VLAs inherit useful priors for perception, object grounding, and instruction understanding, and then attach policy heads, action experts, or action-token interfaces for robot control~\citep{zitkovich2023rt,kim2024openvla,black2024pi_0,intelligence2025pi,pertsch2025fast,cheang2025gr,bjorck2025gr00t,bai2025qwen3,beyer2024paligemma}. Recent VLA progress has also been driven by scaling robot data and improving continuous action generation. Foundation policies collect trajectories across diverse embodiments, tasks, and environments to improve generalization~\citep{wu2026pragmatic,yang2026abot,jiang2025galaxea}, while related systems study cross-embodiment learning, real-time execution, and generalist manipulation policies~\citep{zheng2025x,cen2025rynnvla,cai2026xiaomi,community2026starvla}. For low-level control, flow matching has become a practical formulation because it can model multimodal action distributions while generating smooth action chunks~\citep{lipman2022flow,black2024pi_0}. From a modeling perspective, however, most VLA policies are still trained mainly as direct observation-and-language-to-action mappings, so their supervision for how the physical scene evolves under robot intervention is indirect compared with video-based world modeling.

Among existing VLA systems, DeMaVLA~\citep{su2026demavla} provides the closest real-world reference for our study because it combines a strong VLA policy with a household deformable manipulation protocol. We therefore use DeMaVLA not merely as another baseline, but as a matched VLA reference for DSWAM: the comparison uses the same robot platform, pretraining data, post-training data, task definitions, and success criteria. This alignment is important because comparisons between WAM and VLA policies can otherwise be dominated by differences in data scale, embodiment, or evaluation protocol. Under this matched setting, the real-world evaluation more directly compares VLA-style policy modeling with WAM-style video-co-trained execution.

\subsection{World Action Models}

World Action Models (WAMs) address a complementary limitation of direct VLA policies by using predictive world modeling as supervision for robot action learning. Instead of only mapping current observations to actions, WAMs learn how visual states evolve under robot intervention through future video prediction, video-action co-training, or joint modeling of observations and actions. This provides dense temporal supervision for physical dynamics and can improve manipulation policies on tasks where contact, deformation, and state evolution matter. Recent WAMs have explored several forms of video-based action learning: DreamZero~\citep{ye2026world} builds on a pretrained video diffusion backbone and jointly predicts future videos and continuous actions, Cosmos Policy~\citep{kim2026cosmos} fine-tunes a video foundation model into a robot policy that predicts action chunks, future states, and values, Motus~\citep{bi2025motus} integrates understanding, video generation, and action experts, and LingBot-VA~\citep{li2026causal} studies causal video-action modeling with closed-loop rollout and autoregressive diffusion. Other WAMs move toward richer world representations or more action-centered execution. X-WAM~\citep{guo2026xwam} extends WAMs from 2D video prediction to 4D world-action modeling with multi-view RGB-D futures, while GigaWorld-Policy~\citep{ye2026gigaworld} and Fast-WAM~\citep{yuan2026fastwam} show that policies can retain the benefit of world-modeling supervision while reducing or removing explicit future-video generation at inference.

Despite this progress, most existing WAMs are designed primarily as execution policies. This is a good fit for physically grounded manipulation, but it leaves task-level semantic organization under-specified when household tasks involve coarse instructions that need to be decomposed into ordered executable subtasks. DSWAM addresses this gap through a dual-system design: System 1 remains a WAM executor trained for world-aware physical execution, while System 2 is an optional vision-language subtask planner that supplies explicit decomposition only when it is useful. This preserves the execution strength of WAMs for atomic or already reliable tasks, while adding a language-level planning pathway for complex multi-step commands.

\subsection{Efficient Real-World WAM Deployment}

Deploying WAMs on real robots raises a practical tension: video-based world modeling provides useful physical supervision, but explicit future-video generation can be expensive on the robot control path. Many WAMs follow an imagine-then-execute design in which future observations are generated through denoising, autoregressive decoding, or latent rollout before actions are produced. Such designs can provide visual foresight, but they also increase inference latency and may make closed-loop control depend on imperfect generated futures. Recent work has therefore explored both model-level and system-level acceleration. DreamZero~\citep{ye2026world} uses asynchronous closed-loop execution, decoupled video-action denoising schedules, caching, quantization, and low-level optimization. LingBot-VA~\citep{li2026causal} runs action prediction and robot execution in parallel while feeding real observations back into the rollout. X-WAM~\citep{guo2026xwam} reduces denoising cost with asynchronous noise sampling, and MotuBrain~\citep{motubrain2026advanced} combines action-only inference with denoising-step reduction, compilation, quantization, and caching. Complementary to these approaches, GigaWorld-Policy~\citep{ye2026gigaworld} and Fast-WAM~\citep{yuan2026fastwam} suggest that the main benefit of WAMs can come from training-time world-modeling supervision rather than test-time future imagination.

DSWAM follows this deployment-oriented direction. The executor is trained with action prediction and video co-training, but directly predicts continuous action chunks at inference time without explicit future video generation. We further integrate RTC, asynchronous execution, and TensorRT acceleration so that policy queries do not block robot control. This keeps the WAM executor practical for real-robot control while preserving training-time world-modeling supervision.

\section{The DSWAM Model}
\label{sec:method}

\subsection{Overview}

\begin{figure*}[t]
    \centering
    \includegraphics[width=\textwidth]{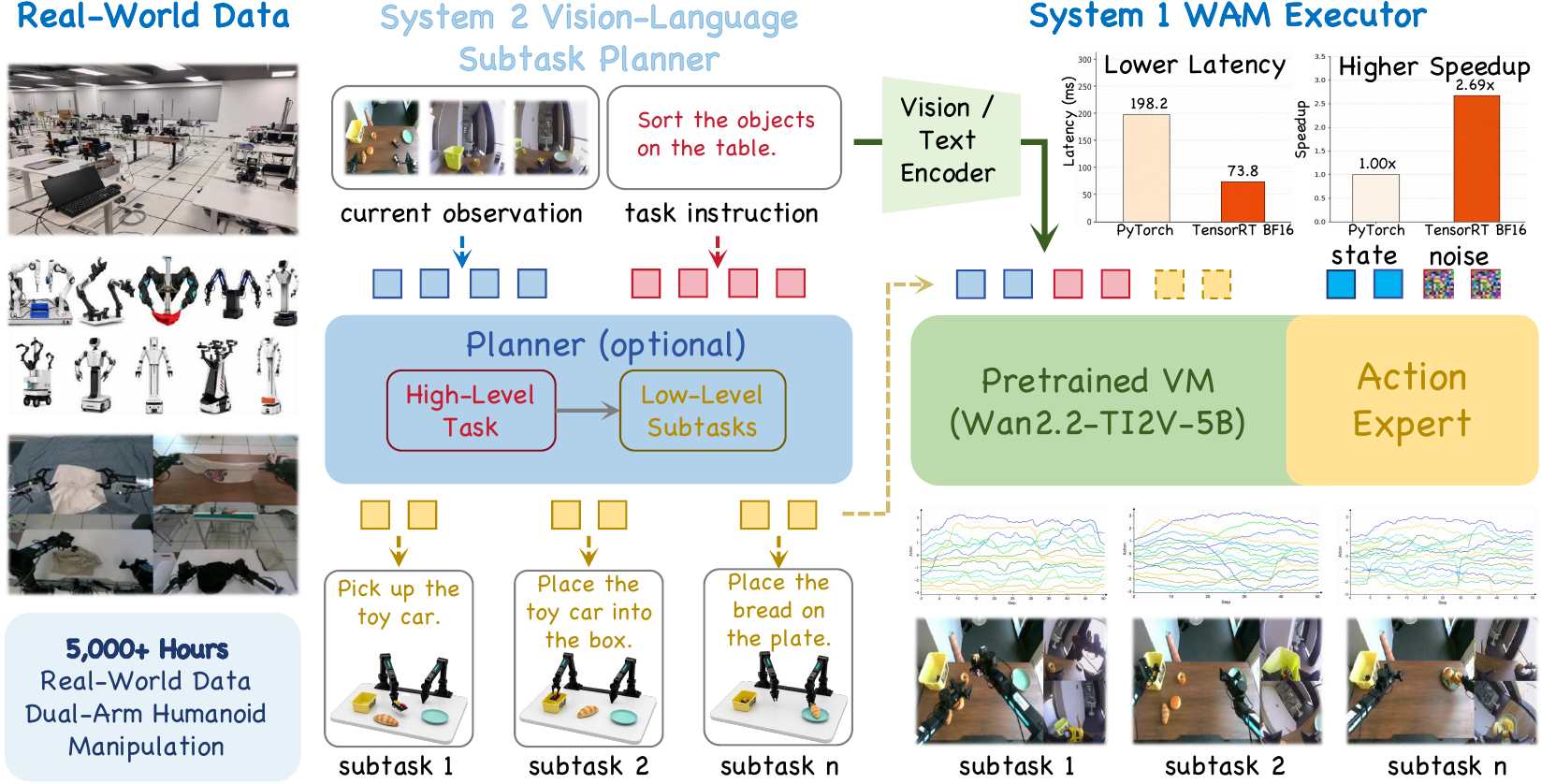}
    \caption{\textbf{Overview of DSWAM.}
    Large-scale real-world data supports scalable training. System 2 is an optional vision-language planner that decomposes coarse instructions into executable subtasks, while System 1 is the default WAM executor that uses a pretrained video model and an action expert to predict action chunks from observations, language conditioning, and proprioception. Video co-training provides training-time world-modeling supervision, while inference avoids explicit future-video generation and uses TensorRT acceleration for efficient deployment.}
    \label{fig:method_overview}
\end{figure*}

DSWAM is a dual-system framework for fine-grained robot manipulation, as shown in Fig.~\ref{fig:method_overview}. The framework combines large-scale real-world robot training, an optional high-level planner, and an efficient WAM executor. System 1 is the default execution path: it conditions on current multimodal observations, language, and proprioception to predict action chunks through a video-model-based WAM executor and an action expert. System 2 is an optional vision-language subtask planner that is invoked only when a command benefits from explicit decomposition. When the planner is not activated, DSWAM conditions System 1 directly on the original prompt and runs the WAM policy without an additional planning step. When the planner is activated, System 2 uses short-term visual history and the global task prompt to decompose a coarse multi-step command into fine-grained executable subtasks, which are then passed to System 1 as language conditions. The executor is trained with action prediction and video co-training, but inference directly predicts actions without explicit future-video generation. This execution-first design keeps planning out of the control path when it is unnecessary, while TensorRT acceleration and asynchronous chunk execution make the same policy practical for real-world deployment.

At timestep $t$, the robot observes multi-view RGB images, a language instruction or subtask, and proprioceptive state:
\begin{equation}
    \mathbf{O}_t = [\mathbf{I}_t^1, \mathbf{I}_t^2, \mathbf{I}_t^3, \ell_t, \mathbf{q}_t],
\end{equation}
where $\mathbf{I}_t^{1,2,3}$ are camera observations, $\ell_t$ denotes either the original instruction or the current System 2 subtask, and $\mathbf{q}_t$ is the robot state. The execution policy predicts an action chunk
\begin{equation}
    \mathbf{A}_t = [\mathbf{a}_t, \mathbf{a}_{t+1}, \dots, \mathbf{a}_{t+H-1}],
\end{equation}
where $H$ is the action horizon and each $\mathbf{a}_t$ is a continuous dual-arm control command.

\subsection{Optional System 2 High-Level Planner}

Some real-world household instructions are coarse enough to require intermediate decisions before execution. For example, a command may imply searching for an object, moving it into a reachable configuration, spreading it, and then executing a task-specific manipulation routine. Existing WAM executors can be strong at physical manipulation, but they typically do not expose an explicit language-level mechanism for deciding which subtask should be executed next. We therefore instantiate System 2 as a multimodal high-level planner based on a Rynnbrain4B-style vision-language model~\citep{cen2025rynnvla}.

Given a global task prompt $p$, System 2 observes a short-term visual history $\mathbf{X}_t$ containing the most recent $T=5$ frames sampled at 1 Hz:
\begin{equation}
    \mathbf{X}_t = \{x_{t-4}, x_{t-3}, x_{t-2}, x_{t-1}, x_t\}, \quad x_i \in \mathbb{R}^{H \times W \times 3}.
\end{equation}
The planner predicts the next subtask-level instruction:
\begin{equation}
    s_t = \mathcal{P}_{\phi}(\mathbf{X}_t, p),
\end{equation}
where $s_t$ is a fine-grained executable command used as the language condition for System 1. When the planner is not activated, DSWAM bypasses System 2 and uses the original prompt directly as the System 1 language condition, i.e., $s_t=p$.

\paragraph{Subtask boundary supervision.}
We construct System 2 training data from robot trajectory videos. Each video is segmented into subtask-level intervals using a pretrained vision-language annotator, yielding subtask labels $s_k$ and temporal boundaries $\tau_k$. Within each segment, frames are sampled at 1 FPS and grouped into sliding windows
\begin{equation}
    \mathbf{W}_i = \{x_i, x_{i+1}, x_{i+2}, x_{i+3}, x_{i+4}\}.
\end{equation}
The supervision target is transition-aware:
\begin{equation}
    y_i =
    \begin{cases}
        s_k, & \mathbf{W}_i \text{ lies inside subtask } k,\\
        s_{k+1}, & \mathbf{W}_i \text{ is the final window of nonterminal subtask } k,\\
        \texttt{done}, & \mathbf{W}_i \text{ is the final window of the terminal subtask}.
    \end{cases}
\end{equation}
This teaches System 2 not only to recognize the current subtask, but also to emit the next instruction near a subtask boundary. The planner is optimized with an autoregressive language modeling objective:
\begin{equation}
    \mathcal{L}_{\mathrm{plan}}(\phi) =
    -\mathbb{E}_{(\mathbf{X}_t,p,y_t)}
    \left[\log P_{\phi}(y_t \mid \mathbf{X}_t, p)\right].
\end{equation}
\paragraph{Coupling with System 1.}
When System 2 is enabled during deployment, the planner and System 1 executor communicate through a synchronized execution protocol. The WAM executor sends a batch of five recent frames to the planner every $\Delta t=2$s:
\begin{equation}
    s_t = \mathcal{P}_{\phi}(\mathbf{X}^{\mathrm{WAM}}_t, p), \quad \Delta t = 2\mathrm{s}.
\end{equation}
The returned subtask $s_t$ conditions the WAM executor until the next planner update. Although System 2 is trained with a fixed 1 FPS sampling regime, we find it robust to the variable-frequency visual inputs produced by real robot execution. In WAM-only mode, the same executor instead conditions directly on the original instruction. This design decouples semantic decomposition from physical execution: System 2 chooses what should be done next only when decomposition is useful, while System 1 determines how to execute the current instruction or subtask through closed-loop action generation.

\subsection{System 1 WAM Executor}

\subsubsection{World Action Model Policy}

The WAM executor is designed to retain the representation benefits of video-based world modeling without paying the inference cost of explicit future video generation. The policy encodes the current multimodal context into latent world features:
\begin{equation}
    \mathbf{z}_t = f_{\theta}(\mathbf{I}_t^{1:3}, \ell_t, \mathbf{q}_t),
\end{equation}
and predicts action chunks from this representation:
\begin{equation}
    p_{\theta}(\mathbf{A}_t \mid \mathbf{O}_t) = p_{\theta}(\mathbf{A}_t \mid \mathbf{z}_t).
\end{equation}
Here $\mathbf{O}_t$ denotes the current multimodal observation context, including multi-view images, the active language condition, and proprioception. During inference, $\mathbf{z}_t$ is produced by a single forward pass over this context. No future frames are sampled, denoised, or decoded. This direct-action interface follows the efficient WAM direction suggested by Fast-WAM~\citep{yuan2026fastwam}, while preserving a training signal that encourages the visual backbone to encode physical dynamics.

The action branch predicts continuous action chunks with conditional flow matching. This is well suited to bimanual manipulation because it represents smooth action trajectories without discretizing robot control into language tokens. The action expert is conditioned on latent world features, proprioception, and the current subtask instruction, allowing the same execution policy to handle different fine-grained manipulation commands.

We train the executor with an action objective and an auxiliary video co-training objective under a shared flow-matching formulation. Let $\mathbf{y}$ denote either an action chunk $\mathbf{A}_t$ or future latent visual tokens $\mathbf{V}_{t+1:t+T}$. For a flow time $\tau \in (0,1)$ and Gaussian noise $\boldsymbol{\epsilon}$, we construct
\begin{equation}
    \mathbf{y}_{\tau} = (1-\tau)\mathbf{y} + \tau \boldsymbol{\epsilon}.
\end{equation}
The model predicts the velocity field $\boldsymbol{\epsilon}-\mathbf{y}$ with the flow-matching loss
\begin{equation}
    \mathcal{L}_{\mathrm{FM}}(\mathbf{y}) =
    \mathbb{E}_{\mathbf{y},\boldsymbol{\epsilon},\tau}
    \left[
        \left\| v_{\theta}(\mathbf{y}_{\tau}, \tau, \mathbf{O}_t) - (\boldsymbol{\epsilon}-\mathbf{y}) \right\|_2^2
    \right].
\end{equation}
The action loss is
\begin{equation}
    \mathcal{L}_{\mathrm{act}} = \mathcal{L}_{\mathrm{FM}}(\mathbf{A}_t),
\end{equation}
and the video co-training loss is
\begin{equation}
    \mathcal{L}_{\mathrm{vid}} = \mathcal{L}_{\mathrm{FM}}(\mathbf{V}_{t+1:t+T}).
\end{equation}
The overall objective is
\begin{equation}
    \mathcal{L} = \mathcal{L}_{\mathrm{act}} + \lambda_{\mathrm{vid}}\mathcal{L}_{\mathrm{vid}}.
\end{equation}
During training, future visual tokens provide world-modeling supervision, but the action tokens are prevented from using future visual tokens as privileged information. This keeps the action policy aligned with the inference-time setting, where only current observations are available.

\subsubsection{Real-time Chunking and Deployment}

For real-world deployment, DSWAM uses inference-time RTC~\citep{black2026real} and asynchronous execution to decouple policy queries from the robot control loop. This is a deployment-time mechanism: we do not use training-time action conditioning or a suffix-only training loss, and the executor is trained to predict complete action chunks with the objectives described above. At inference time, a policy worker predicts new chunks from the latest observations while the controller continues executing the current chunk, so policy latency does not block low-level control.

We further compile the transformer-heavy execution path with TensorRT to reduce runtime latency. Instead of exporting one monolithic graph, DSWAM uses two engines. The visual-context engine builds the per-layer cache from the video latent, diffusion timestep, language/proprioceptive context, and visual attention mask. The action-denoising engine consumes action latents, the same context, the cached visual state, and the joint video-action attention mask to predict one denoising update for the action chunk. This split preserves the public serving interface: preprocessing, text encoding, VAE handling, normalization, and policy wrapping remain unchanged, while the dominant visual and action transformer computation is delegated to TensorRT.

In our real-robot experiments, we use the BF16 TensorRT path because it preserves close end-to-end agreement with the PyTorch reference while substantially reducing policy-query latency. Together with RTC and no-future-generation WAM inference, TensorRT acceleration keeps DSWAM practical for low-latency robot control.

\section{Experiments}
\label{sec:experiment}

We organize the experiments from simulation to real-world deployment. We first evaluate broad bimanual manipulation capability on RoboTwin 2.0, then use the matched DeMaVLA folding protocol to compare WAM-style execution with VLA policies under aligned robot platform, training data, task definitions, and evaluation criteria. DSWAM runs in WAM-only mode in this folding benchmark, with the System 2 planner disabled, so the comparison focuses on policy execution capability. We further study whether optional System 2 subtask supervision improves tasks that benefit from decomposition, and profile TensorRT and asynchronous RTC to evaluate real-robot inference efficiency.

\subsection{Experimental Setup}

\paragraph{RoboTwin 2.0 protocol.}
We evaluate on RoboTwin 2.0~\citep{chen2025robotwin}, which contains 50 bimanual manipulation tasks under clean and randomized settings. The clean setting uses fixed initial configurations, while the randomized setting varies object poses and scene layouts. We report per-task success rates and average success rate across the full task suite.

\paragraph{Matched real-world folding protocol.}
To provide a controlled real-robot comparison with VLA policies, we follow the household folding protocol introduced by DeMaVLA~\citep{su2026demavla}. The benchmark uses an ALOHA-style dual-arm platform and evaluates four deformable manipulation tasks: folding a shirt, folding a skirt, folding pants, and folding a towel. For each garment category, we evaluate two physical garment instances with 10 trials per instance, resulting in 20 trials per category. One instance is used as the easier setting, while the other is used as the hard setting; the hard garment is either not included in the data collection for that category or is physically aged and worn. Each trial covers the full household procedure: the garment is randomly dropped into a basket, the robot retrieves it, places it on the table, spreads it into a foldable configuration, and completes the target folding routine. In this folding benchmark, DSWAM is deliberately evaluated in WAM-only mode: the System 2 planner is not started, and the executor conditions on the original folding instruction throughout the rollout. This keeps the benchmark a fair comparison of policy execution capability under matched robot platform, pretraining data, post-training data, task protocol, and success criteria.

\paragraph{System 2 sorting protocol.}
For the optional System 2 study, we use a real-world tabletop sorting task with the coarse instruction \textit{Sort objects on the table}. The subtask-supervised setting decomposes this command into two executable instructions: \textit{pick up the toy car and place it into the box} and \textit{pick up the bread and place it on the plate}. We randomize the initial object configurations and compare WAM executors trained with the raw coarse instruction against executors trained with the subtask-level instructions.

\paragraph{Metrics.}
For RoboTwin 2.0, we report Success Rate (SR) under clean and randomized settings. For the real-world folding benchmark, we report SR and average Completion Time, aggregated over the easy and hard garment instances in each category. A real-world folding trial is successful if the robot completes the full procedure and produces a valid folded configuration within the time limit. A trial is counted as a failure if the robot exceeds the time limit or the garment falls off the table. Completion time reflects both task efficiency and failure frequency, since failed trials are counted using the timeout duration, following the DeMaVLA protocol. For the System 2 sorting study, we report SR and the average number of execution mistakes per rollout.

\paragraph{Compared methods.}
For RoboTwin 2.0, we compare against representative VLA and WAM baselines: $\pi_0$~\citep{black2024pi_0}, $\pi_{0.5}$~\citep{intelligence2025pi_}, DeMaVLA~\citep{su2026demavla}, Fast-WAM~\citep{yuan2026fastwam}, and Motus~\citep{bi2025motus}. For real-world folding, we compare with $\pi_0$ and DeMaVLA under the same evaluation protocol. DSWAM uses the executor policy directly, and the System 2 planner is not launched in this comparison, making the evaluation focus on policy design and execution capability rather than differences in robot embodiment, data collection, or additional high-level planning.

\subsection{RoboTwin 2.0 Simulation Benchmark}

Table~\ref{tab:robotwin_full} reports the per-task RoboTwin 2.0 results. This benchmark complements the real-world folding experiments by covering a broader set of simulated bimanual manipulation skills. The clean setting tests the nominal task distribution, while the randomized setting evaluates robustness to changed object poses and scene layouts.

DSWAM reaches 92.38\% average SR under the clean setting and 91.90\% under the randomized setting, achieving the best average performance in both settings. The gains are not limited to comparisons with WAM-style baselines: DSWAM also outperforms the VLA baselines by a clear margin. Compared with DeMaVLA, the strongest VLA baseline in this table, DSWAM improves average SR by 3.96\% in the clean setting and 5.12\% in the randomized setting. Compared with $\pi_{0.5}$, the gains are 9.64\% and 15.14\%, respectively. Relative to Fast-WAM, DSWAM further improves average SR by 0.50\% in the clean setting and 0.12\% in the randomized setting, showing that the proposed executor preserves broad task coverage while improving robustness under randomization.

\begin{table}[p]
\centering
\caption{\raggedright \textit{Evaluation on RoboTwin 2.0 Simulation Benchmark}. Success rates are reported under clean and randomized settings across 50 bimanual manipulation tasks.}
\label{tab:robotwin_full}
\setlength{\tabcolsep}{3.0pt}
\resizebox{0.93\textwidth}{!}{%
\begin{tabular}{l cc cc cc cc cc cc}
\toprule
\multirow{2}{*}{\textbf{Simulation Task}} &
\multicolumn{2}{c}{$\pi_0$} & \multicolumn{2}{c}{$\pi_{0.5}$} & \multicolumn{2}{c}{DeMaVLA} & \multicolumn{2}{c}{Motus} & \multicolumn{2}{c}{Fast-WAM} & \multicolumn{2}{c}{DSWAM} \\
\cmidrule(lr){2-3} \cmidrule(lr){4-5} \cmidrule(lr){6-7} \cmidrule(lr){8-9} \cmidrule(lr){10-11} \cmidrule(lr){12-13}
 & Clean & Rand. & Clean & Rand. & Clean & Rand. & Clean & Rand. & Clean & Rand. & Clean & Rand. \\
\midrule
\textit{Adjust Bottle} & 99\% & 95\% & 100\% & 99\% & 99\% & 100\% & 89\% & 93\% & 100\% & 100\% & 100\% & 100\% \\
\textit{Beat Block Hammer} & 79\% & 84\% & 96\% & 93\% & 79\% & 85\% & 95\% & 88\% & 99\% & 97\% & 100\% & 98\% \\
\textit{Blocks Ranking RGB} & 80\% & 63\% & 92\% & 85\% & 95\% & 95\% & 99\% & 97\% & 100\% & 100\% & 100\% & 100\% \\
\textit{Blocks Ranking Size} & 14\% & 5\% & 49\% & 26\% & 72\% & 68\% & 75\% & 63\% & 94\% & 98\% & 84\% & 85\% \\
\textit{Click Alarmclock} & 77\% & 68\% & 98\% & 89\% & 98\% & 100\% & 100\% & 100\% & 100\% & 100\% & 100\% & 100\% \\
\textit{Click Bell} & 71\% & 48\% & 99\% & 66\% & 96\% & 98\% & 100\% & 100\% & 100\% & 100\% & 100\% & 100\% \\
\textit{Dump Bin Bigbin} & 88\% & 83\% & 92\% & 97\% & 91\% & 94\% & 95\% & 91\% & 97\% & 96\% & 88\% & 93\% \\
\textit{Grab Roller} & 98\% & 94\% & 100\% & 100\% & 100\% & 100\% & 100\% & 100\% & 100\% & 100\% & 100\% & 100\% \\
\textit{Handover Block} & 47\% & 31\% & 66\% & 57\% & 93\% & 83\% & 86\% & 73\% & 95\% & 81\% & 96\% & 93\% \\
\textit{Handover Mic} & 97\% & 97\% & 98\% & 97\% & 95\% & 96\% & 78\% & 63\% & 99\% & 100\% & 96\% & 94\% \\
\textit{Hanging Mug} & 14\% & 11\% & 18\% & 17\% & 46\% & 35\% & 38\% & 38\% & 58\% & 62\% & 61\% & 51\% \\
\textit{Lift Pot} & 80\% & 72\% & 96\% & 85\% & 97\% & 94\% & 96\% & 99\% & 100\% & 100\% & 100\% & 99\% \\
\textit{Move Can Pot} & 68\% & 48\% & 51\% & 55\% & 93\% & 79\% & 34\% & 74\% & 90\% & 88\% & 97\% & 93\% \\
\textit{Move Pillbottle Pad} & 67\% & 46\% & 84\% & 61\% & 86\% & 86\% & 93\% & 96\% & 100\% & 99\% & 99\% & 100\% \\
\textit{Move Playingcard Away} & 74\% & 65\% & 96\% & 84\% & 94\% & 90\% & 100\% & 96\% & 100\% & 100\% & 99\% & 100\% \\
\textit{Move Stapler Pad} & 41\% & 24\% & 56\% & 42\% & 83\% & 76\% & 83\% & 85\% & 77\% & 64\% & 78\% & 72\% \\
\textit{Open Laptop} & 71\% & 81\% & 90\% & 96\% & 98\% & 100\% & 95\% & 91\% & 98\% & 100\% & 99\% & 99\% \\
\textit{Open Microwave} & 4\% & 32\% & 34\% & 77\% & 83\% & 72\% & 95\% & 91\% & 62\% & 45\% & 60\% & 43\% \\
\textit{Pick Diverse Bottles} & 69\% & 31\% & 81\% & 71\% & 58\% & 75\% & 90\% & 91\% & 80\% & 85\% & 99\% & 100\% \\
\textit{Pick Dual Bottles} & 59\% & 37\% & 93\% & 63\% & 89\% & 75\% & 96\% & 90\% & 100\% & 96\% & 100\% & 100\% \\
\textit{Place A2B Left} & 43\% & 47\% & 87\% & 82\% & 92\% & 95\% & 88\% & 79\% & 95\% & 93\% & 94\% & 88\% \\
\textit{Place A2B Right} & 39\% & 34\% & 87\% & 84\% & 90\% & 87\% & 91\% & 87\% & 93\% & 99\% & 93\% & 94\% \\
\textit{Place Bread Basket} & 62\% & 46\% & 77\% & 64\% & 83\% & 84\% & 91\% & 94\% & 91\% & 93\% & 96\% & 97\% \\
\textit{Place Bread Skillet} & 66\% & 49\% & 85\% & 66\% & 94\% & 83\% & 86\% & 83\% & 90\% & 93\% & 89\% & 93\% \\
\textit{Place Burger Fries} & 81\% & 76\% & 94\% & 87\% & 96\% & 95\% & 98\% & 98\% & 96\% & 99\% & 96\% & 98\% \\
\textit{Place Can Basket} & 55\% & 46\% & 62\% & 62\% & 87\% & 77\% & 81\% & 76\% & 71\% & 69\% & 64\% & 68\% \\
\textit{Place Cans Plasticbox} & 63\% & 45\% & 94\% & 84\% & 89\% & 92\% & 98\% & 94\% & 99\% & 96\% & 100\% & 100\% \\
\textit{Place Container Plate} & 97\% & 92\% & 99\% & 95\% & 96\% & 98\% & 98\% & 99\% & 96\% & 100\% & 98\% & 99\% \\
\textit{Place Dual Shoes} & 59\% & 51\% & 75\% & 75\% & 96\% & 94\% & 93\% & 87\% & 94\% & 88\% & 85\% & 88\% \\
\textit{Place Empty Cup} & 91\% & 85\% & 100\% & 99\% & 99\% & 99\% & 99\% & 98\% & 100\% & 100\% & 100\% & 100\% \\
\textit{Place Fan} & 66\% & 71\% & 87\% & 85\% & 94\% & 91\% & 91\% & 87\% & 96\% & 96\% & 95\% & 94\% \\
\textit{Place Mouse Pad} & 20\% & 20\% & 60\% & 39\% & 75\% & 78\% & 66\% & 68\% & 83\% & 89\% & 93\% & 88\% \\
\textit{Place Object Basket} & 67\% & 70\% & 80\% & 76\% & 84\% & 66\% & 81\% & 87\% & 89\% & 88\% & 82\% & 87\% \\
\textit{Place Object Scale} & 57\% & 52\% & 86\% & 80\% & 90\% & 89\% & 88\% & 85\% & 90\% & 97\% & 98\% & 95\% \\
\textit{Place Object Stand} & 82\% & 68\% & 91\% & 85\% & 93\% & 92\% & 98\% & 97\% & 90\% & 94\% & 96\% & 94\% \\
\textit{Place Phone Stand} & 49\% & 53\% & 81\% & 81\% & 95\% & 90\% & 87\% & 86\% & 97\% & 99\% & 95\% & 98\% \\
\textit{Place Shoe} & 76\% & 76\% & 92\% & 93\% & 100\% & 100\% & 99\% & 97\% & 96\% & 99\% & 95\% & 93\% \\
\textit{Press Stapler} & 44\% & 37\% & 87\% & 83\% & 96\% & 97\% & 93\% & 98\% & 90\% & 97\% & 87\% & 92\% \\
\textit{Put Bottles Dustbin} & 65\% & 56\% & 84\% & 79\% & 88\% & 85\% & 81\% & 79\% & 95\% & 90\% & 92\% & 94\% \\
\textit{Put Object Cabinet} & 73\% & 60\% & 80\% & 79\% & 92\% & 85\% & 88\% & 71\% & 94\% & 89\% & 92\% & 93\% \\
\textit{Rotate QRcode} & 74\% & 70\% & 89\% & 87\% & 95\% & 86\% & 89\% & 73\% & 93\% & 89\% & 92\% & 89\% \\
\textit{Scan Object} & 55\% & 42\% & 72\% & 65\% & 79\% & 83\% & 67\% & 66\% & 89\% & 92\% & 91\% & 94\% \\
\textit{Shake Bottle Horizontally} & 98\% & 92\% & 99\% & 99\% & 100\% & 99\% & 100\% & 98\% & 100\% & 100\% & 100\% & 100\% \\
\textit{Shake Bottle} & 94\% & 91\% & 99\% & 97\% & 100\% & 100\% & 100\% & 97\% & 100\% & 100\% & 99\% & 100\% \\
\textit{Stack Blocks Three} & 72\% & 52\% & 91\% & 76\% & 97\% & 95\% & 91\% & 95\% & 95\% & 97\% & 99\% & 96\% \\
\textit{Stack Blocks Two} & 93\% & 79\% & 97\% & 100\% & 100\% & 98\% & 100\% & 98\% & 100\% & 100\% & 100\% & 100\% \\
\textit{Stack Bowls Three} & 77\% & 75\% & 77\% & 71\% & 83\% & 80\% & 79\% & 87\% & 80\% & 81\% & 83\% & 86\% \\
\textit{Stack Bowls Two} & 94\% & 95\% & 95\% & 96\% & 98\% & 96\% & 98\% & 98\% & 92\% & 98\% & 93\% & 97\% \\
\textit{Stamp Seal} & 46\% & 33\% & 79\% & 55\% & 77\% & 83\% & 93\% & 92\% & 90\% & 94\% & 89\% & 93\% \\
\textit{Turn Switch} & 41\% & 42\% & 62\% & 54\% & 18\% & 41\% & 84\% & 78\% & 61\% & 59\% & 77\% & 67\% \\
\midrule
\textbf{Average (\%)} & 65.92 & 58.40 & 82.74 & 76.76 & 88.42 & 86.78 & 88.66 & 87.02 & 91.88 & 91.78 & \textbf{92.38} & \textbf{91.90} \\
\bottomrule
\end{tabular}%
}
\vspace{-5pt}
\end{table}

\subsection{Matched Real-world Folding Benchmark}

Table~\ref{tab:real_world_benchmark} shows the matched real-world comparison. DSWAM achieves the best average result, reaching 96.3\% SR and $1'44''$ average completion time. Relative to DeMaVLA, DSWAM improves average SR by 3.8 percentage points and reduces average completion time by 34 seconds. Relative to $\pi_0$, it improves average SR by 20.0 percentage points and reduces average completion time by 42 seconds.

\clearpage

\begin{center}
\centering
\captionsetup{type=table,hypcap=false}
\caption{Real-world evaluation on the household folding benchmark. DSWAM is evaluated in WAM-only mode with the System 2 planner disabled. Each garment category uses two physical instances with 10 trials per instance, covering one easier setting and one hard setting. SR denotes success rate, and Time denotes average completion time.}
\label{tab:real_world_benchmark}
\begin{tabular}{l cc cc cc cc cc}
\toprule
\multirow{2}{*}{\textbf{Method}} &
\multicolumn{2}{c}{\textbf{Shirt}} &
\multicolumn{2}{c}{\textbf{Skirt}} &
\multicolumn{2}{c}{\textbf{Pant}} &
\multicolumn{2}{c}{\textbf{Towel}} &
\multicolumn{2}{c}{\textbf{Average}} \\
\cmidrule(lr){2-3} \cmidrule(lr){4-5} \cmidrule(lr){6-7} \cmidrule(lr){8-9} \cmidrule(lr){10-11}
 & SR & Time & SR & Time & SR & Time & SR & Time & SR & Time \\
\midrule
$\pi_0$ & 90.0\% & $1'55''$ & 95.0\% & $1'03''$ & 65.0\% & $3'01''$ & 55.0\% & $3'44''$ & 76.3\% & $2'26''$ \\
DeMaVLA & 95.0\% & $2'15''$ & 100.0\% & $1'30''$ & 75.0\% & $3'01''$ & 100.0\% & $2'26''$ & 92.5\% & $2'18''$ \\
DSWAM & \textbf{95.0\%} & $\bm{2'14''}$ & \textbf{100.0\%} & $\bm{0'58''}$ & \textbf{90.0\%} & $\bm{2'19''}$ & \textbf{100.0\%} & $\bm{1'27''}$ & \textbf{96.3\%} & $\bm{1'44''}$ \\
\bottomrule
\end{tabular}
\end{center}

\begin{center}
\centering
\captionsetup{type=table,hypcap=false}
\caption{Effect of subtask-level instruction supervision on the real-world sorting task. Subtask supervision improves success rate and reduces execution mistakes for the coarse instruction \textit{Sort objects on the table}.}
\label{tab:system2_subtask_supervision}
\begin{tabular}{l c c c}
\toprule
\textbf{Instruction Setting} & \textbf{Training Steps} & \textbf{SR} & \textbf{Mistakes / Rollout} \\
\midrule
Raw task instruction & 6000 & 71.4\% & 3.75 \\
Raw task instruction & 18000 & 80.0\% & 3.30 \\
Subtask-level instruction & 6000 & \textbf{100.0\%} & \textbf{1.00} \\
Subtask-level instruction & 18000 & \textbf{100.0\%} & \textbf{0.30} \\
\bottomrule
\end{tabular}
\end{center}

The improvement is most visible on pants, where DSWAM increases SR from 75.0\% to 90.0\% relative to DeMaVLA while reducing completion time from $3'01''$ to $2'19''$. On skirts and towels, DSWAM maintains 100.0\% SR and substantially shortens the rollout. On shirts, it matches DeMaVLA's 95.0\% SR while being slightly faster. Because the protocol matches the robot, data, task definitions, and evaluation criteria, and because DSWAM keeps the System 2 planner disabled in this benchmark, these gains provide controlled evidence for the strength of WAM-style policy execution in real-world deformable manipulation, rather than benefiting from differences in embodiment, data, planning, or evaluation setup.

\subsection{Optional System 2 Subtask Supervision Study}

We next isolate the effect of the optional System 2 branch using the sorting protocol described above. This study does not assume that every instruction should be decomposed; instead, it asks whether subtask-level supervision helps when a coarse command has a clear task structure.

Table~\ref{tab:system2_subtask_supervision} shows that subtask-level supervision improves execution stability in this setting. Averaged over the reported evaluations, the raw sorting instruction achieves 75.7\% SR with 3.53 mistakes per rollout, while subtask-level instructions reach 100.0\% SR and reduce mistakes to 0.65 per rollout. These results support the optional use of System 2: when a command needs decomposition, explicit subtask context can guide the WAM executor and reduce error accumulation; when the instruction is already atomic, the planner branch does not need to be invoked.

\FloatBarrier

\subsection{Efficiency and Deployment}

Finally, we evaluate whether the executor can run under real-robot timing constraints. DSWAM uses video co-training as a training signal, but at inference time the action path predicts action chunks directly from the current observation and language condition without explicit future-video generation. We combine this inference path with BF16 TensorRT acceleration and asynchronous RTC for deployment. As shown in Table~\ref{tab:trt_latency}, BF16 TensorRT reduces warmed end-to-end policy latency from 198.2 ms in PyTorch to 73.8 ms on an NVIDIA GeForce RTX 5090 with CUDA 12.9 and TensorRT 10.16.1, giving a 2.69$\times$ speedup while maintaining close action agreement with the PyTorch reference.

Table~\ref{tab:async_rtc_success} reports real-robot rollouts with \textit{synchronous TensorRT} and \textit{asynchronous TensorRT+RTC} on the easy garment instances. This rollout study is used to isolate the deployment mechanism; therefore, its absolute completion times should be interpreted separately from the full folding benchmark in Table~\ref{tab:real_world_benchmark}, which averages over both easy and hard garments. \textit{Asynchronous TensorRT+RTC} preserves 100\% success on shirts and improves pants from 70\% to 100\%, while reducing average success time on both tasks.

\clearpage

\begin{center}
\centering
\captionsetup{type=table,hypcap=false}
\caption{TensorRT inference profiling. Latency is warmed end-to-end policy inference with batch size 1. Action agreement is measured against the PyTorch reference over six seeds.}
\label{tab:trt_latency}
\begin{tabular}{l c c c c}
\toprule
\textbf{Execution Path} & \textbf{Latency (ms)} & \textbf{Speedup} & \textbf{Max Rel. Err.} & \textbf{Cosine Sim.} \\
\midrule
PyTorch & 198.2 & 1.00$\times$ & -- & -- \\
TensorRT BF16 & \textbf{73.8} & \textbf{2.69}$\times$ & \textbf{0.0106} & \textbf{0.99977} \\
\bottomrule
\end{tabular}
\end{center}

\begin{center}
\centering
\captionsetup{type=table,hypcap=false}
\caption{Real-robot rollout results for \textit{synchronous TensorRT} and \textit{asynchronous TensorRT+RTC} on easy garment instances. These trials evaluate deployment efficiency and are not directly comparable to the full easy+hard folding benchmark times.}
\label{tab:async_rtc_success}
\begin{tabular}{l l c c c}
\toprule
\textbf{Setting} & \textbf{Task} & \textbf{Trials} & \textbf{SR} & \textbf{Avg. Success Time} \\
\midrule
Synchronous TensorRT & Shirt & 10 & 100\% & $1'47''$ \\
Synchronous TensorRT & Pants & 10 & 70\% & $1'50''$ \\
Asynchronous TensorRT+RTC & Shirt & 10 & \textbf{100\%} & $\bm{1'28''}$ \\
Asynchronous TensorRT+RTC & Pants & 10 & \textbf{100\%} & $\bm{1'08''}$ \\
\bottomrule
\end{tabular}
\end{center}

These results indicate that efficient deployment is beneficial for folding, where continuous bimanual manipulation and contact-rich adjustment require frequent action updates; decoupling policy inference from robot control can improve execution stability without changing the trained policy.

\FloatBarrier

\section{Conclusion}
\label{sec:conclusion}

We presented DSWAM, a dual-system World Action Model that decouples semantic task organization from physically grounded robot execution. DSWAM uses a WAM executor as the default System 1 policy and invokes an optional System 2 planner only when a coarse instruction benefits from subtask decomposition, allowing atomic commands to remain direct while still supporting structured multi-step household tasks. The executor uses video co-training to learn world-aware representations, but predicts action chunks directly at inference time without explicit future-video generation. Through a matched real-world comparison with VLA policies under the same robot platform, data, task protocol, and evaluation criteria, DSWAM provides controlled evidence that WAM-style execution is a strong foundation for real-world deformable manipulation; this comparison uses the WAM executor directly without planner assistance, so the result reflects policy execution capability rather than additional high-level planning. The System 2 study further shows that subtask-level supervision can improve stability when decomposition is useful, while the deployment study shows that TensorRT acceleration and asynchronous RTC make the policy practical for efficient on-robot execution. Overall, DSWAM points to a scalable design for robot foundation policies: preserve the physical execution strengths of WAMs, add explicit task decomposition only when needed, and keep inference efficient enough for real-world manipulation.

\clearpage
\bibliographystyle{plainnat}
\bibliography{main}

\end{document}